\documentclass{article}

\usepackage[final, main]{Template/_}

\usepackage[utf8]{inputenc} 
\usepackage[T1]{fontenc}    
\usepackage{hyperref}       
\usepackage{url}            
\usepackage{booktabs}       
\usepackage{amsfonts}       
\usepackage{nicefrac}       
\usepackage{microtype}      
\usepackage{xcolor}         
\usepackage{amsmath,graphicx}
\usepackage{mathtools}
\usepackage{float}
\usepackage{wrapfig}
\usepackage{caption}
\usepackage{floatflt}
\usepackage{cleveref}  
\usepackage{siunitx}   
\usepackage{float}            

\newcommand{\CDelta}{\mathbf{\Delta}}

\let\citep\cite
\let\citet\cite
\title{Physics Steering: Causal Control of Cross-Domain Concepts in a Physics Foundation Model}

\author{
  Rio Alexa Fear \\
  University of Cambridge\\
  \texttt{raf74@cam.ac.uk} \\
  \And
  Payel Mukhopadhyay \\
  University of Cambridge\\
  \texttt{pm858@cam.ac.uk} \\
  \AND
  Michael McCabe \\
  NYU \& Simons Foundation\\
  \texttt{mmccabe@simonsfoundation.org} \\
  \And
  Alberto Bietti \\
  Flatiron Institute \\
  \texttt{abietti@flatironinstitute.org} \\
  \And
  Miles Cranmer \\
  University of Cambridge\\
  \texttt{mc2473@cam.ac.uk} \\
  \AND
  The PolymathicAI Collaboration
}

\begin{document}

\maketitle

\begin{abstract}
Foundation models trained on text and images have been shown to develop abstract internal features that align with human concepts and that can be directly manipulated via activation steering to alter model behaviour. Whether scientific foundation models learn similarly abstract and domain-general representations has remained an open question. Inspired by recent work identifying single directions in activation space that control complex behaviours in large language models, we show that Walrus, a large physics foundation model, learns linearly steerable representations of physical phenomena. By computing the difference between activations representing contrasting physical regimes, we identify single directions in activation space that correspond to vorticity, diffusion, and even temporal progression. We find that injecting these concept directions back into the model during inference enables fine-grained causal control: vortices can be induced or removed, diffusion enhanced or suppressed, and simulations sped up or slowed down. Moreover, directions derived from one physical system also steer the behaviour of other physical systems, suggesting that they are domain-general. These results provide early evidence that scientific foundation models can learn general representations of physical principles, and further support for the Linear Representation Hypothesis.
\end{abstract}

\section{Introduction}
Recent advances in the field of interpretability have enhanced our comprehension of how foundation models function. Methods such as probing \citep{vig2020causal, wang2022interpretability} and Sparse Autoencoders (SAEs) \citep{bricken2023monosemanticity, cunningham2023sparse}, primarily designed for large language models (LLMs), have uncovered that these models often form internal representations, or hidden features, that closely resemble human concepts \citep{devlin2018bert, brown2020language, he2021masked}. A wide range of features from descriptive nouns \cite{elhage2021framework}, to abstract meta-concepts, and more \cite{templeton2024scaling, lindsey2025biology} are well documented in the literature. Recent studies have even indicated that intricate behaviours such as refusal can be mediated by a single direction in activation space \citep{arditi2024refusal}. Importantly, these features are not merely correlational; interventions such as activation steering demonstrate that they have a causal influence on model behaviour \citep{turner2023activation, zou2023representation, li2023inference}. 

The scientific community is increasingly leveraging large-scale models developed on extensive datasets across diverse domains, including chemistry \citep{chithrananda2020chemberta, bran2023chemcrow}, astronomy \citep{nguyen2023astrollama, leung2023astronomical,Parker_2024}, climate science \citep{nguyen2023climax}, and healthcare \citep{jiang2023health}. However, while there has been rapid progress in interpretability research for LLMs, the internal representations of foundation models trained on scientific data remain largely unexplored. A key open question is whether --- in a manner similar to LLMs --- simulation models form interpretable representations which align with fundamental physical laws and principles, or if they depend on superficial correlations and patterns in the data.

This paper begins to tackle these questions by applying interpretability techniques adapted from LLM studies to Walrus \citep{mccabe2025walruscrossdomainfoundationmodel}, a new state-of-the-art transformer model pretrained on The Well \citep{ohana2024well}, a large and varied collection of PDE simulations. Our research aims to determine if Walrus learns interpretable internal representations of physical phenomena and whether these representations can be causally manipulated. Inspired by the method of identifying single direction "concept vectors" described by \citep{arditi2024refusal, zou2023representation}, we employ a modified version of the technique to determine directions in the model's activation space which correspond to specific physical concepts. We inject these concept vectors during the model's forward pass to achieve activation steering \citep{turner2023activation} and thereby assess their causal impact on the resulting simulations.

Our contributions include:
\begin{itemize}
    \item A methodology for extracting interpretable physical concept features from transformer-based physics models.
    \item We compute single-direction "delta" tensors between activations from contrasting physical regimes.
    \item We show that intervention along these directions causally steers model predictions in interpretable ways.
    \item We provide evidence that concept features are transferable between unrelated physical systems, suggesting that neural networks learn transferable abstract concepts across different physics domains.
\end{itemize}

\subsection{Background}
\paragraph{The Physics Foundation Model.} Walrus is a large vision transformer \citep{vaswani2017attention, ho2019axial} based foundation model designed for spatiotemporal surrogate modelling of physical systems described by PDEs. Walrus has been pretrained on a large range of complex and diverse datasets present in the Well collection \citep{ohana2024well}. It builds upon similar physics foundation model approaches introduced by \citep{hao2024dpot, herde2024poseidon, cao2025vicon, mccabe2024multiple}. In short, Walrus is trained autoregressively to predict the next state of a physical system given a sequence of previous states. A key aspect of this pretraining is aiming to learn broadly useful representations of physical dynamics and facilitate transfer learning.

\paragraph{The Well.} \citep{ohana2024well} is a large-scale (15TB) benchmark dataset comprising 16 distinct numerical simulations curated in collaboration with domain experts. It spans diverse fields including fluid dynamics (e.g., Rayleigh-Bénard convection, Shear Flow, Magneto-hydrodynamics), astrophysics (e.g., Supernovae, Post-neutron star mergers), acoustic scattering, and even biological systems; amongst other things. 

The data is provided as sequences of snapshots on uniform grids, and for each simulation includes multiple trajectories with varying initial conditions or physical parameters. The Well provides the diverse, high-quality data necessary for the physics foundation model to learn representations that generalize across physical domains and provides a challenging benchmark for evaluating generalization and transfer in scientific ML \citep{takamoto2022pdebench}. The Well also serves as the testbed for our interpretability investigations.

\subsection{Interpretability}
Mechanistic interpretability aims to reverse engineer neural networks into human-understandable algorithms \citep{olah2020zoom, nanda2022mechanistic}. Several key interpretability hypotheses underpin this work, these are briefly covered below.

\paragraph{Linear representation hypothesis.} posits that features (i.e., concepts) are represented linearly as directions in a models activation space \citep{elhage2022toy, arora2018linear, park2024linear}. 

\paragraph{Polysemanticity.} refers to the theory that deep learning models can represent more features than the dimensionality of their activation space would suggest. Models achieve this by assigning multiple, potentially unrelated, features to a single neuron (\emph{polysemanticity}) and representing features in non-orthogonal directions (\emph{superposition}) \citep{scherlis2023polysemanticity, elhage2022toy, henighan2023superposition, arora2018linear}. 

An unfortunate side effect of polysemanticity is that it complicates interpretation, as individual polysemantic neurons are generally not easily interpretable. Various techniques (e.g. SAEs) aim to address this by creating new representations of internal activations where neurons and features have a 1 to 1 relationship, resulting in \emph{monosemantic} features \citep{bricken2023monosemanticity, templeton2024scaling}.

It should be noted that a monosemantic feature is not necessarily a feature which makes sense to a human being. For instance, one can imagine an LLM learning a monosemantic feature based purely on complex correlations between tokens which bears no relationship whatsoever to any human concept. After all, a feature is just a reusable, statistically independent component of a dataset that a model happens to find useful. Yet recent interpretability research has demonstrated that LLMs (and vision transformers) \textit{do} indeed learn a multitude of human-interpretable features. \citep{meng2022locating, wang2022interpretability}.

Many researchers argue that concept-based features emerge as a side effect of structure which is intrinsic to the training data. Language possesses inherent syntactic and semantic hierarchies that reflect human concepts, therefore text data provides a rich, human-understandable symbolic structure that models can pick up directly from the text. It is thus perhaps unsurprising that language data should lend itself to the formation of meaningful high-level abstractions and concept features in LLMs.

Numerical physics data, on the other hand, lacks an explicit concept structure. Instead, this structure can only arise indirectly through the abstraction of underlying governing rules, which a model must first infer. Therefore, for a foundation model trained on physics data, there is less \textit{a priori} reason to assume its internal representations will correspond to human concepts

\paragraph{Activation steering.} Beyond passive observation, interpretability aims to achieve causal understanding. Activation steering is a causal intervention technique where a precomputed vector, representing a concept, is added to the models activations at a specific layer during a forward pass. If the concept vector is meaningful and the intervention is successful, the models output will change in a manner which is consistent with the concept. This serves to test the causal link between activation directions and the models behaviour \citep{turner2023activation, zou2023representation, li2023inference, templeton2024scaling}. Activation steering has been used to control stylistic attributes, factual recall, and more. 

\paragraph{Single Direction Steering.} Our work draws on the approach outlined by \citet{arditi2024refusal}, which showed that complex behaviours in LLMs, such as refusal behaviour, can be identified with a single direction in activation space. This direction can be found by the computation of concept \emph{deltas}, that is, by finding differences between model activations for different inputs (e.g., toxic vs. non-toxic text), one can identify directions in activation space that correspond to specific concepts. These directions can then be added or subtracted from activations during inference to steer the model’s behaviour. 

\section{Methodology}
Our methodology consists of four main steps: (1) selection of contrasting simulation files representing two distinct physical regimes; (2) extraction of activations from forward passes of Walrus across several examples from each regime; (3) calculation of "delta" concept directions; and (4) injection of concept directions to steer model outputs. 

\begin{figure}[!ht]
    \centering
    \includegraphics[width=0.95\linewidth]{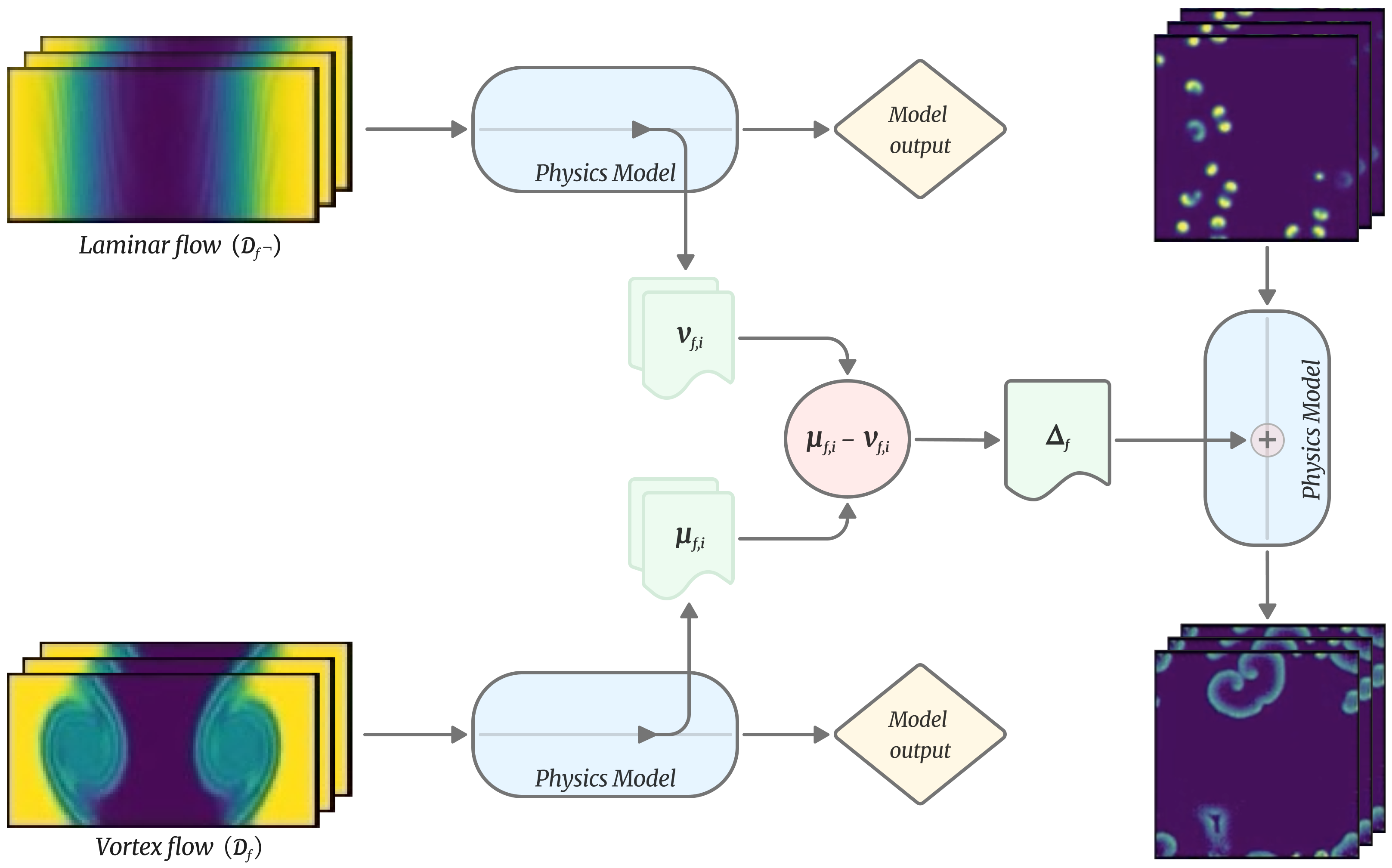}
    \caption{Schematic illustration of methodology. Activations are first extracted from the physics model during forward passes over input segments that exhibit physical feature \smash{$f$}, yielding activations $\boldsymbol{\mu}_{f,i}$, and from segments lacking the feature, \smash{${\neg}f$}, yielding $\boldsymbol{\nu}_{f,i}$. The difference between these activations, $\CDelta_f$, is then injected back into Walrus during inference to steer future results.}
    \label{fig:schematic}
\end{figure}

We investigate whether single-direction activation interventions, termed "delta steering", can be used to understand and control the internal representations of physical phenomena within Walrus, a recently released state-of-the-art physics foundation model. We therefore test the hypothesis that physical concepts are linearly represented in the latent space of physics foundation models. Let $\mathbf{a}$ denote the activation tensor for a particular transformer block of a physics foundation model. We seek to identify direction $\CDelta_f$ in activation space such that the intervention $\mathbf{a} \rightarrow \mathbf{a} + \alpha \CDelta_f$ for scalar $\alpha > 0$ causally steers the model's predictions toward a desired physical feature $f$. The methodology employed consists of four primary steps, adapted from techniques used in LLM interpretability.

\begin{enumerate}

\item \textbf{Selection of Contrasting Simulation Files:} We create two groups of simulations taken from The Well such that the groups represent two distinct regimes of one physical system, with the difference between them being some physical feature which has visually distinguishable macro-scale effects, complex dynamics emerging from micro-dynamics, and the existence analogous structures across different phenomena to enable transferability studies. To meet these criteria, we focused our initial investigations on vorticity within the Shear Flow dataset, chosen for its well-understood physics and distinct visual features. 

\item \textbf{Activation Extraction: } Activations were extracted from Walrus during forward passes over selected input segments from the simulation trajectories. PyTorch \citep{paszke2019pytorch} hooks were used to capture the activations from the final processor block immediately before the decoder  ($blocks.39$), hypothesised to be the most likely to contain abstract representations of physical dynamics \citep{geva2021transformer}. However, noting the observations in \citep{skean2025layer}, we suspect that features in intermediate-to-late layers might achieve similar effects, and exploring this layer-dependence would be an interesting direction for future work. The model was run in rollout mode, processing windowed segments of consecutive timesteps. For each input, we extract the activation tensor $\mathbf{a} \in \mathcal{A} \subseteq \mathbb{R}^{T\times C \times W \times H}$, where $\mathcal{A}$ is the activation space, $T$ is the sequence length, $C$ the channel/feature dimension, and $W$, $H$ are spatial dimensions (width and height). 

\item \textbf{Calculation of Concept Directions:} The saved tensors were averaged across each group resulting in an average laminar flow tensor and an average vortex flow tensor. The "delta tensor", or concept direction, were then computed by taking the difference between the two averaged activation tensors.

Let $\mathcal{D}_{f}$ denote the dataset of activation tensors extracted from input segments that exhibit physical feature $f$, and let $\mathcal{D}_{\neg{f}}$ denote the dataset from input segments that lack feature $f$ or exhibit the opposite of feature $f$. To identify the direction corresponding to the physical feature $f$, we first normalize and then average the activations.

For each activation position $i = (t, w, h) \in \mathcal{I}$, where $\mathcal{I}$ is the set of all activation positions in the model's representation and these positions correspond to spatiotemporal locations in the physical simulation, we normalize the activations:

\begin{equation}
    \hat{\mathbf{a}}_i = \frac{\mathbf{a}_i - \bar{\mathbf{a}}_i}{\sigma_i}
\end{equation}

where $\bar{\mathbf{a}}_i$ and $\sigma_i$ are the mean and standard deviation across the training data at position $i$. We then compute the mean normalized activations for each dataset:

\begin{equation}
{\boldsymbol{\mu}}_{f,i} \coloneqq \frac{1}{|\mathcal{D}_{f}^{\text{train}}|}
\sum_{\mathbf{a} \in \mathcal{D}_{f}^{\text{train}}}
\hat{\mathbf{a}}_i ,
\quad
\:\:
{\boldsymbol{\nu}}_{f,i} \coloneqq \frac{1}{|\mathcal{D}_{\neg{f}}^{\text{train}}|}
\sum_{\mathbf{a} \in \mathcal{D}_{\neg{f}}^{\text{train}}}
\hat{\mathbf{a}}_i
\end{equation}

and we compute the concept direction as the difference between averaged activations:

\begin{equation}
    \Delta_{f,i} \coloneqq \boldsymbol{\mu}_{f,i} - \boldsymbol{\nu}_{f,i}
\end{equation}

yielding the full concept direction tensor $\CDelta_f \in \mathcal{A}$. This direction is interpreted as encoding the concept of physical feature $f$ in activation space.

For cross-domain transfer experiments where spatial structures may not align between different physical systems, we also compute a spatially-averaged concept direction:

\begin{equation}
    \overline{\CDelta}_f \coloneqq \frac{1}{|\mathcal{I}|} \sum_{i \in \mathcal{I}} \Delta_{f,i}
\end{equation}

This spatially-averaged direction $\overline{\CDelta}_f \in \mathbb{R}^C$ preserves only the channel-wise concept information. 

\item \textbf{Activation Steering (Injection of Concept Directions):} To test the causal influence of these concept directions, they were injected back into the model during inference. Using a forward hook at the same target layer, the original activations $\mathbf{a}$ were modified by addition of the concept direction. The modified activations $\mathbf{a}'$ were calculated with steering function $\mathbf{s}: \mathcal{A} \times \mathcal{A} \times \mathbb{R} \rightarrow \mathcal{A}$ where  $\mathbf{a}, \CDelta_f \in \mathcal{A}$ and $\alpha \in \mathbb{R}$:
\begin{equation}
\mathbf{s}(\mathbf{a}, \CDelta_f, \alpha) \coloneq \mathbf{a} + \alpha\|\mathbf{a}\|^2\frac{\CDelta_f}{\|\CDelta_f\|^2}
\label{eq:injection}
\end{equation}

where $\alpha$ is a scaling factor. The output was then renormalised to preserve the original norm of $\mathbf{a}$. This intervention was applied across all tokens and time steps.
\end{enumerate}

\section{Experiments}
\subsection{Progressive Suppression of a Physical Feature}
The most straightforward method of testing the causal influence of the concept direction is to suppress the physical feature $a$ in the output simulation by means of subtraction in \cref{eq:injection}. However, for ease of interpretation, the choice of the physical feature $a$ is crucial.

\paragraph{Result.} The effect of negative activation steering was visually striking. Whereas the unmodified simulation displayed two prominent vortical structures, the steered simulations showed a progressive suppression of these features with increasing $\alpha$. The flow was instead transformed into a smooth, parallel state characteristic of a laminar regime. The successful laminarisation of the flow is an encouraging first sign that our method can precisely target and remove specific complex phenomena from a simulation.

\begin{figure}[!ht]
    \centering
    \includegraphics[width=0.8\linewidth]{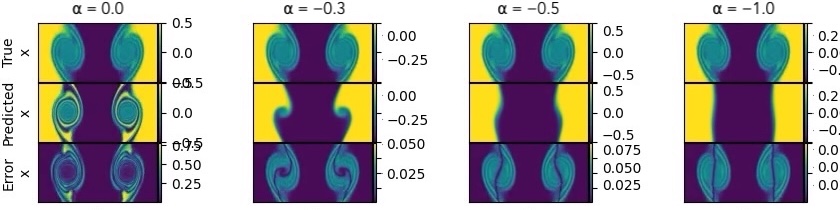}
    \caption{Negative \(\CDelta_{\text{vortex}}\) injection into shear flow vortex regime, for $\alpha$ values of 0, 0.3, 0.5 and 1.0. Frame: 64.}
    \label{fig:vortex_suppress_shear}
\end{figure}

\subsection{Continuous induction of a Physical Feature} 
A natural subsequent question to ask is whether the opposite intervention will be similarly effective --- can the addition of the same concept vector to the activations at layer $l$ give rise to the associated physical feature in the output simulation? 

Successful inducement of a feature is a notably higher bar to pass than simple suppression because Walrus predicts token deltas at each time step, rather than the entire state of each token, so it is conceivable that the feature suppression intervention may not truly be targeting a feature representing a physical characteristic. It may instead be setting the prediction deltas to zero for a range of tokens, thereby resulting in the initial simulation state (i.e., laminar flow), persisting throughout the model rollout window.

To address this concern we repeated the suppression procedure but with the sign reversed in \cref{eq:injection}.

\paragraph{Result.} Positive injection of the learned vortex direction during inference on shear-flow simulations in the laminar regime reliably induced vortical structures, with the effect scaling with the steering strength $\alpha$: small injections ($\alpha\approx\SIrange{0.1}{0.4}{}$) produced subtle perturbations and incipient rotation, while moderate injections ($\alpha\approx\SIrange{0.4}{0.5}{}$) yielded well-formed vortices. 

This result further validates the physical interpretation of the vortex direction and suggests that the extracted direction may encode a meaningful, controllable physical feature capable of introducing vortex formation into an otherwise laminar regime.

\begin{figure}[!ht]
    \centering
    \includegraphics[width=0.8\linewidth]{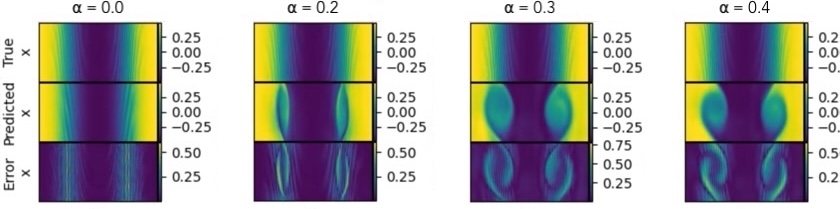}
    \caption{Positive \(\CDelta_{\text{vortex}}\) injection into shear flow laminar regime, for $\alpha$ values of 0, 0.2, 0.3 and 0.4. Frame: 64.}
    \label{fig:vortex_addition_shear}
\end{figure}

The success of the concept induction experiment raises a question: given that vortices are being introduced to a simulation in which they would not normally arise, by what mechanism does Walrus transform a nominally laminar flow to produce a vortex? Is it applying a physically valid initial perturbation and then correctly simulating the ongoing natural evolution? Is it simulating a modified but self-consistent version of the physics? Or is it cosmetically shifting the output to look more like the target concept?

\subsection{Additional Physical Features} 
Given the success of both suppression and induction of the vorticity concept direction we next asked whether an alternative, very different concept can be found? Where a vortex is a localised phenomenon which is defined by its structure, we now aim to isolate a concept direction that represents process-based phenomenon, which is not defined by a specific structure or confined to a particular location: 

\begin{figure}[!ht]
    \centering
    \includegraphics[width=0.25\textwidth]{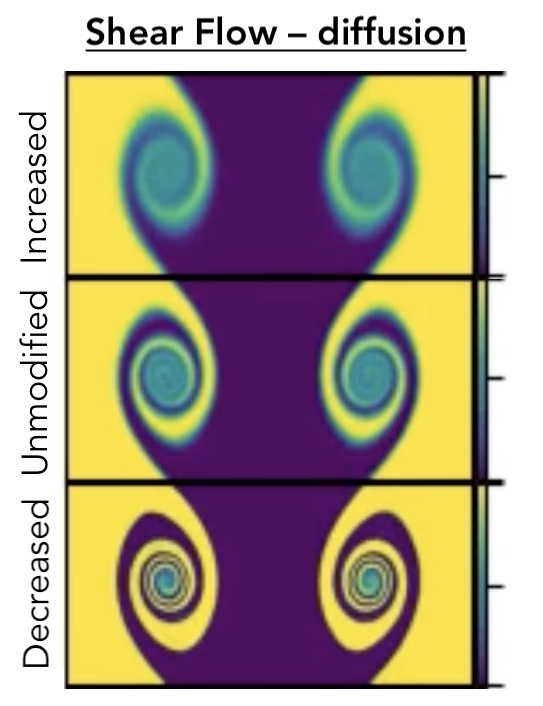}
    \hspace{1cm} 
    \includegraphics[width=0.25\textwidth]{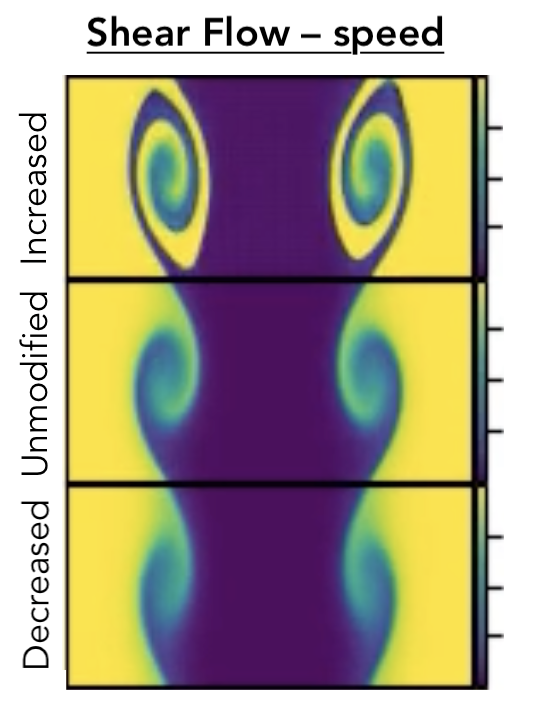}
    \captionsetup{width=0.75\linewidth}
    \caption{On the left tracer fields for $\scriptstyle\CDelta_{\text{diffusion}}$ injection into Shear Flow vortex regime with (top) $\alpha=0.1$ and (bottom) $\alpha=-0.1$. On the right tracer fields for $\scriptstyle\CDelta_{\text{speed}}$ injection into Shear Flow vortex regime with (top) $\alpha=\mathit{0.1}$ and (bottom) $\alpha=\mathit{-0.1}$. Frame(left): 30, Frame(right): 24.}
    \label{fig:shear_additional_concepts}
\end{figure}

\subsubsection{Diffusion} 
Using the same Shear Flow simulation and the same extraction and injection methodology we computed the diffusion delta direction as the difference between the averaged activations for several high molecular diffusion and low molecular diffusion Shear Flow data files – that is, two groups of Shear Flow simulations with identical Reynolds numbers but different sets of Schmidt numbers.

\paragraph{Result.} We discover analogous results for diffusion phenomena, with the diffusion direction encoding meaningful information about diffusion processes which can be manipulated causally. In \cref{fig:shear_additional_concepts} addition of the diffusion direction presented itself as a more diffuse looking fluid interface, while subtraction led to a more sharply defined interface. In appendix \cref{fig:diffusion_fields} addition also leads to larger, more spread out core pressure minima and y-velocity high/low zones, plus smoother x-velocity gradients; subtraction on the other hand leads to a reduction in the size of the same regions, along with sharper x-velocity gradients.

\subsubsection{Temporal} 
After isolating a structural feature (vorticity) and a process-based one (diffusion), we investigated whether a more fundamental simulation property – its temporal progression – could be similarly controlled. To create a "speed" feature, we used the same extraction and injection methodology on two Shear Flow simulations that were physically identical but sampled at different frame rates. The delta direction was computed as the difference between the mean activations of a high-frame-rate (fast) simulation and a low-frame-rate (slow) one.

\paragraph{Result.} Injecting the "speed" direction with a positive steering coefficient caused the vortex to form much earlier in the rollout window. Conversely, subtracting the direction delayed the formation of the vortex. This can seen by the fact that the vortex in the top right of \cref{fig:shear_additional_concepts} is larger and more well developed compared to the lower right image where the vortex has barely formed by the same video frame.

\subsection{Feature Transfer Between Physical Systems}
A final important question arises regarding the nature and usefulness of the discovered concept directions: Are these concept directions specific to the Shear Flow dataset which was used to derive them, or do they represent a more general physical understanding learned by the physics foundation model? 

We thus tested the transferability of the vorticity and speed features by applying the delta concept injection derived from the Shear Flow datasets to three alternative Well datasets, ordered by increasing dissimilarity from the Shear Flow dataset.
\begin{enumerate}
    \item \textbf{Rayleigh-Bénard Convection}: An alternative fluid dynamics dataset which models fluid heated from below and cooled from above, creating convection patterns.
    \item \textbf{Euler Quadrant}: A second alternative fluid dynamics dataset which seeks to simulate two compressible, inviscid gas species governed by the Euler equations.
    \item \textbf{Gray-Scott Reaction-Diffusion}: An entirely unrelated system outside the field of fluid dynamics. This dataset contains simulations of a chemical reaction-diffusion system which produces various pattern formations, including gliders, spots, spirals, and mazes depending on parameter settings.
\end{enumerate}

For within-domain steering (e.g., shear flow to shear flow), we use the full concept direction tensor $\CDelta_f$ that preserves spatial structure. However, when transferring concept directions between different physical systems, the activation tensors may have different spatial dimensions. To address this, we employ two strategies:

\begin{itemize}
    \item \textbf{Spatial averaging}: Using the spatially-averaged concept direction $\overline{\CDelta}_f$ defined in Equation (5), which preserves only channel-wise information. This approach assumes that the physical concept is encoded primarily in the channel dimensions rather than specific spatial patterns.
    \item \textbf{Spatial alignment}: When spatial dimensions are similar (differing by at most one element), we pad or interpolate to match dimensions, preserving spatial structure. Interpolation and padding produced nearly identical results, so we describe the results below in terms of the inclusion or non-inclusion of spatial dimensions.
\end{itemize}

Our experiments show that spatial averaging generally produces more interpretable and physically consistent results for cross-domain transfer, as it extracts the abstract concept independent of system-specific spatial configurations.

\paragraph{Rayleigh-Bénard Vorticity Transfer.} In the first concept transfer experiment we see a clear illustration of the impact which the presence of spatial dimensions in the steering tensor can have. Across each of the Rayleigh-Bénard results we see that the intervention appears to manifest as moderate changes to convection in the buoyancy field, in addition to comparatively extreme shifts in the pressure field. Two primary observations jump out to the viewer: Firstly, when the spatial dimensions are not included there is an increase in convection (that is, convection patterns  appear earlier and are larger) with positive steering, and a corresponding decrease with negative steering. Secondly, when the spatial dimensions are included the simple positive direction = increase and negative direction = decrease relationship disappears. Instead \textit{both} directions produce an increase in convection in the buoyancy field, along with large high and low pressure zones which appear to be inverted between the two results. 

\paragraph{Euler Vorticity Transfer.} Here we see a more straightforward result: an increase in the size and number of rotational flow features in the positive steering direction, especially at shock interfaces. Conversely, in the negative direction we see a decrease in size and number of rotational flow features. It is interesting to note that the shock interfaces are precisely where one would expect vortices to show up in a physically real scenario.

\begin{figure}[!ht]
    \centering
    \includegraphics[width=0.25\linewidth]{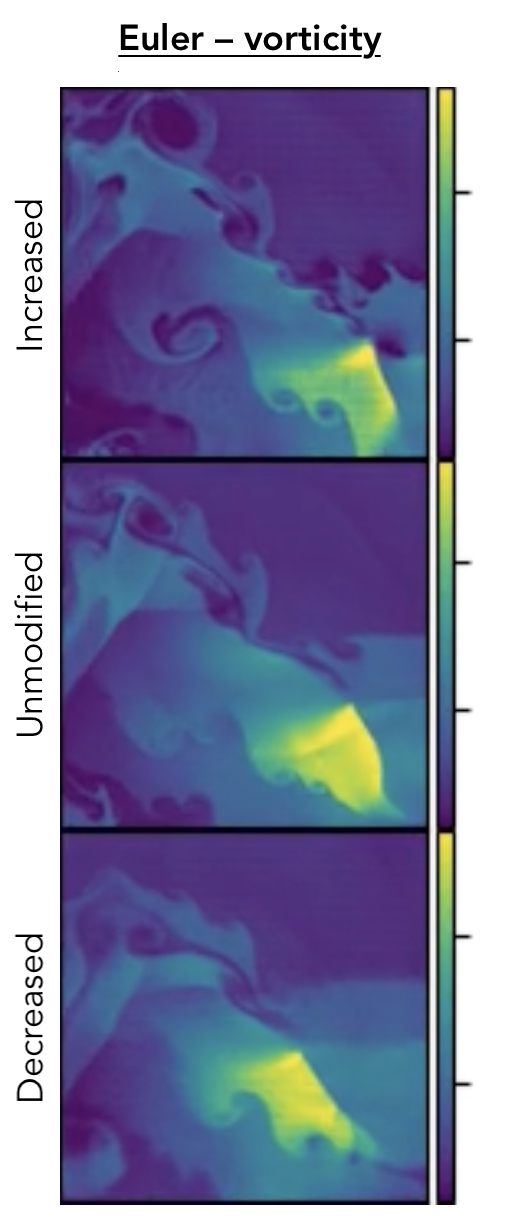}
    \includegraphics[width=0.25\linewidth]{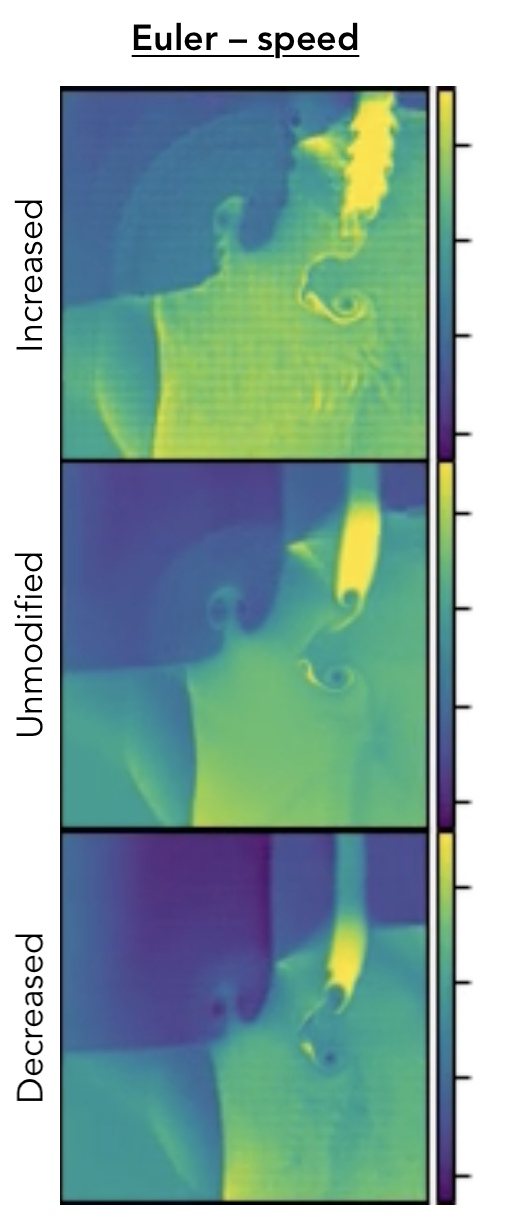}  
    \includegraphics[width=0.25\linewidth]{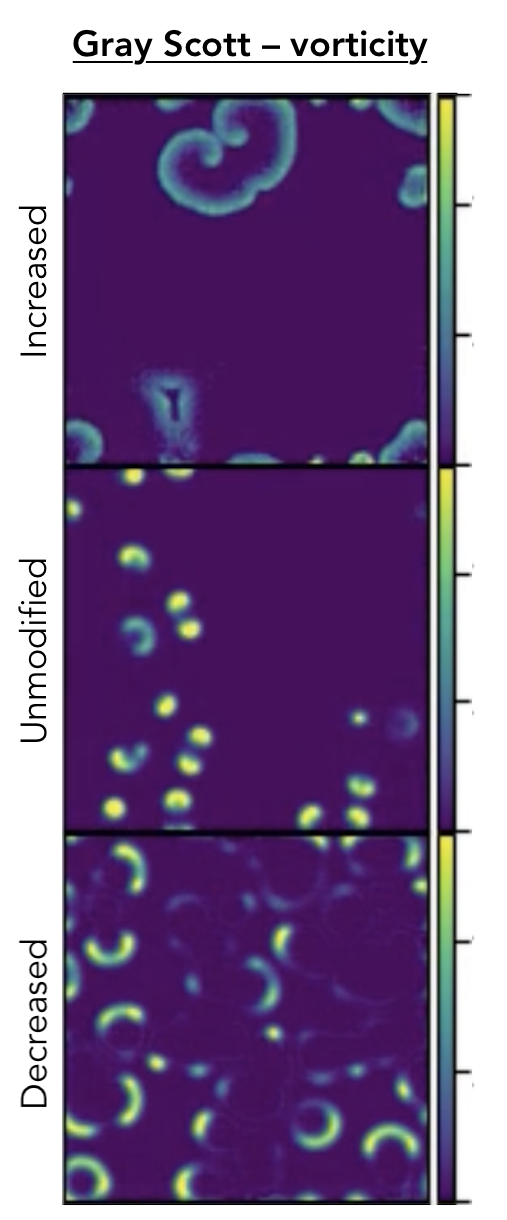}
    \caption{(Left) Density field for $\scriptstyle\overline{\CDelta}_{\text{vortex}}$ injection into Euler quadrants. (Middle) Density field for $\scriptstyle\overline{\CDelta}_{\text{speed}}$ injection into Euler quadrants. (Right) Chemical species B for $\scriptstyle\overline{\CDelta}_{\text{vortex}}$ injection into Gray-Scott reaction diffusion. All computed by averaging over spatial dimensions, with $\alpha=0.1$ (top), $\alpha=0.0$ (middle), and $\alpha=-0.1$ (bottom). Frames: 50 (left), 28 (middle), 48 (right).}
    \label{fig:euler-gray-transfer}
\end{figure}

\paragraph{Euler Speed Transfer.} In the second Euler transfer result it is immediately apparent that the shock fronts move faster with positive steering and slower with negative steering. In \cref{fig:euler-gray-transfer} the effect can most readily be observed by comparing the position of the vertical shock line along the bottom of each of the three images. In the positive (top) image it is further along than the unmodified (middle) image, which in turn is further along than the negative (bottom) image. Another notable observation is that the addition of the speed direction has led to the creation of rotational features along both sides of the thick yellow shock front in the top right of the image.

\paragraph{Gray-Scott Vorticity Transfer.} Of all our results, the most surprising was vorticity steering in the Gray-Scott "gliders" simulation, a physical system which is defined by interactions between two chemical species ("A" and "B") and where the concept of a fluid vortex does not apply. Despite this, we find that positive vortex steering induced the transformation of gliders in the chemical concentration fields into spiral patterns very reminiscent of those normally found in a "spirals" type Gray-Scott Reaction Diffusion system.

\begin{figure}[!ht]
    \centering
    \includegraphics[width=0.75\linewidth]{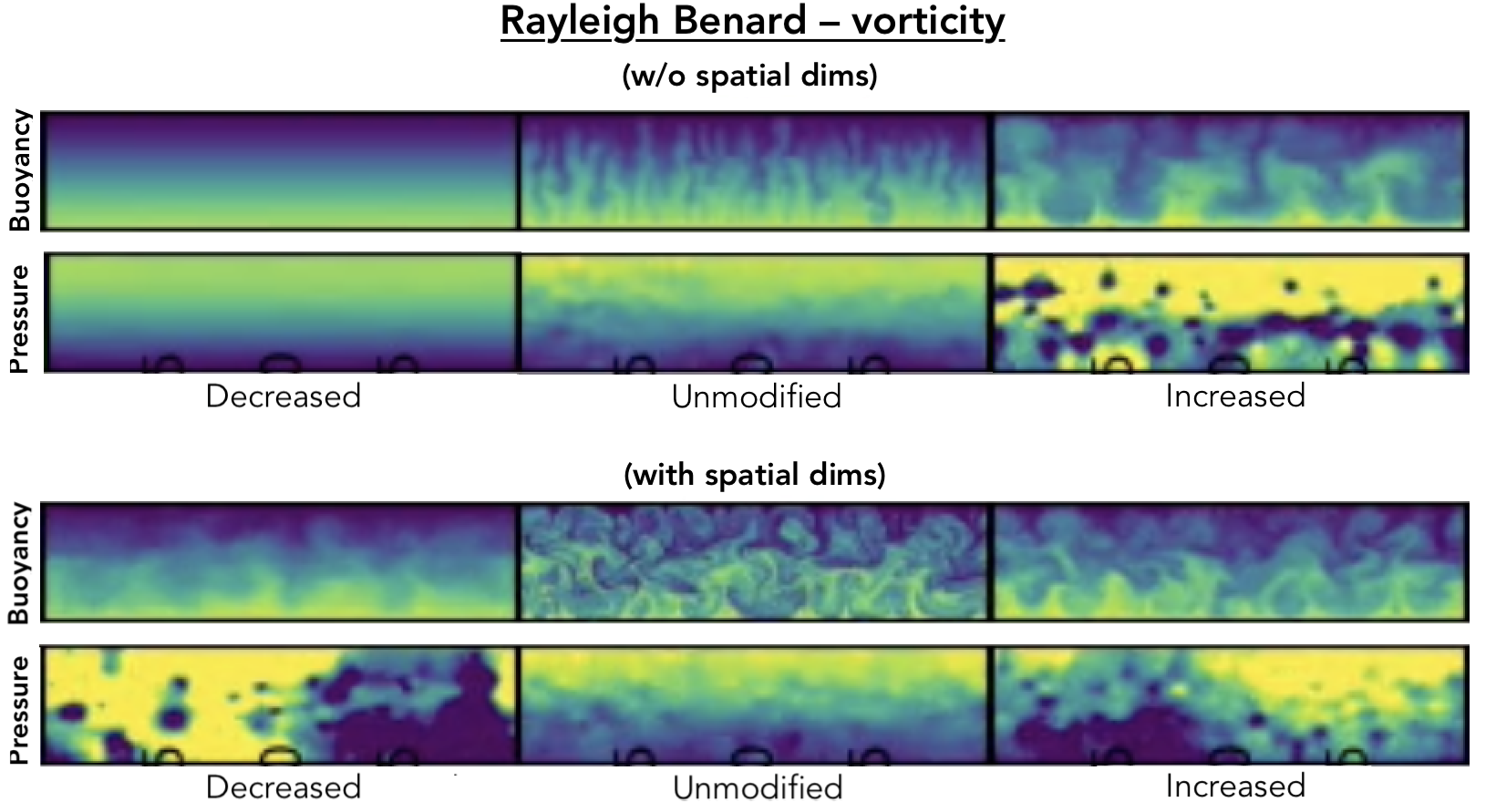}
    \caption{Transfer of $\scriptstyle\CDelta_{\text{vortex}}$ concept injection to Rayleigh-Bénard simulations. Pressure and buoyancy fields for (top) averaging over spatial dimensions: (left) $\alpha=-0.1$, (centre) $\alpha=0.0$, (right) $\alpha=0.1$; (bottom) including spatial dimensions (no averaging): (left) $\alpha=-0.1$, (centre) $\alpha=0.0$, (right) $\alpha=0.1$. Frame(top): 40, Frame(bottom): 50.}
    \label{fig:ray-transfer}
\end{figure}

\vspace{-0.5em}
\section{Discussion}
These experiments show that these concept directions are not merely correlational but have a causal effect on the simulation. Within the Shear Flow dataset, adding the vortex direction induced vortical structures in a laminar flow, while subtracting it suppressed existing vortices.

Interestingly, these concept directions appear to generalise across different physical systems. The vortex direction, derived entirely from Shear Flow simulations, introduced broadly analogous rotational structures when transferred to other fluid dynamics datasets like Rayleigh-Bénard convection and Euler quadrant flows. Most remarkably, when applied to the Gray-Scott reaction-diffusion system—a chemical system where fluid vortices are not physically defined—the same intervention produced spiral patterns. Results which suggest that Walrus may have learned an abstract representation of "rotation" or "spiralling" that transcends any specific physical domain. 

A key open problem --- and a limitation of this work --- is the question of the physicality of the steered results. As a general rule we find that inclusion of spatial dimensions in a transfer steering tensor results in more physically unrealistic results in the secondary visualisation fields, but when those dimensions are averaged over and dropped the results often appear physically plausible. Having said that, it is hard to define what a "reasonable" and "physically plausible" result should actually look like in this context since by its very nature we are aiming to introduce a physical feature into a system where that feature should not naturally be present. Further work is needed to definitively answer the question; for now, we have noted some encouraging observations:
\begin{enumerate}
    \item Steering modifies all fields in the simulation in a physically self-consistent manner. 
    \item The effects of steering tend to show up in physically expected places (e.g. vortices appearing along shock fronts in Euler, gliders transforming into spirals in Gray-Scott; plus more diffuse flow and smoother gradients in the diffusion steering experiment.).
\end{enumerate}
We have also observed that the distance of the simulation's initial conditions from the desired physical regime is important. For example, when performing positive vorticity steering on laminar shear flow with varying Reynolds and Schmidt numbers, we find that the further the initial conditions are from the vortex regime, the harder it will be (i.e. the higher alpha will need to be) to finally force a vortex into the simulation. By the time you succeed (if you do at all), the high alpha will have caused the y velocity and pressure fields to become completely distorted and unphysical. Conversely, if the initial conditions are such that the simulation is on the brink of the vortex regime, only a small nudge will be required, leaving all of the fields looking far more reasonable. 
\vspace{-0.5em}
\section{Conclusion}
Our work demonstrates that interpretability techniques from LLMs can be successfully adapted to scientific foundation models. By calculating the difference in mean activations between contrasting physical regimes we isolate single directions in the latent space of the physics foundation model that correspond to specific physical concepts like vorticity, diffusion, and simulation speed. Injecting these directions during inference provides direct, causal control over the model's predictions, allowing us to manipulate physical behaviours in silico.

These findings provide early evidence that scientific foundation models, much like LLMs, develop abstract, domain-general representations of fundamental concepts. The success of our simple difference-of-means approach suggests that these core physical concepts are represented strongly and linearly in the model's activation space, aligning with the linear representation hypothesis. 

The emergence of steerable, interpretable features in scientific models has significant implications. It increases our confidence that these models are learning genuine physical principles rather than superficial correlations. It opens new avenues for interacting with simulations: we can perform counterfactual exploration ("what if this flow were more diffuse?"), correct simulation errors in real-time, and audit a model's understanding of physics by testing its response to targeted interventions.

\section*{Acknowledgments}
We thank Schmidt Sciences and the Simons Foundation for their support. We are grateful to Shirley Ho, Francois Lanusse, and the rest of the PolymathicAI team for valuable discussions and feedback. We also thank Neel Nanda, Rich Turner, and Max Welling for valuable discussions.

\bibliographystyle{plain}
\bibliography{references}

\appendix
\section{Additional Plots}
\subsection{Spatial Dimensions for Within-Domain Steering}
\begin{figure}[H]
    \centering
    \includegraphics[width=0.8\linewidth]{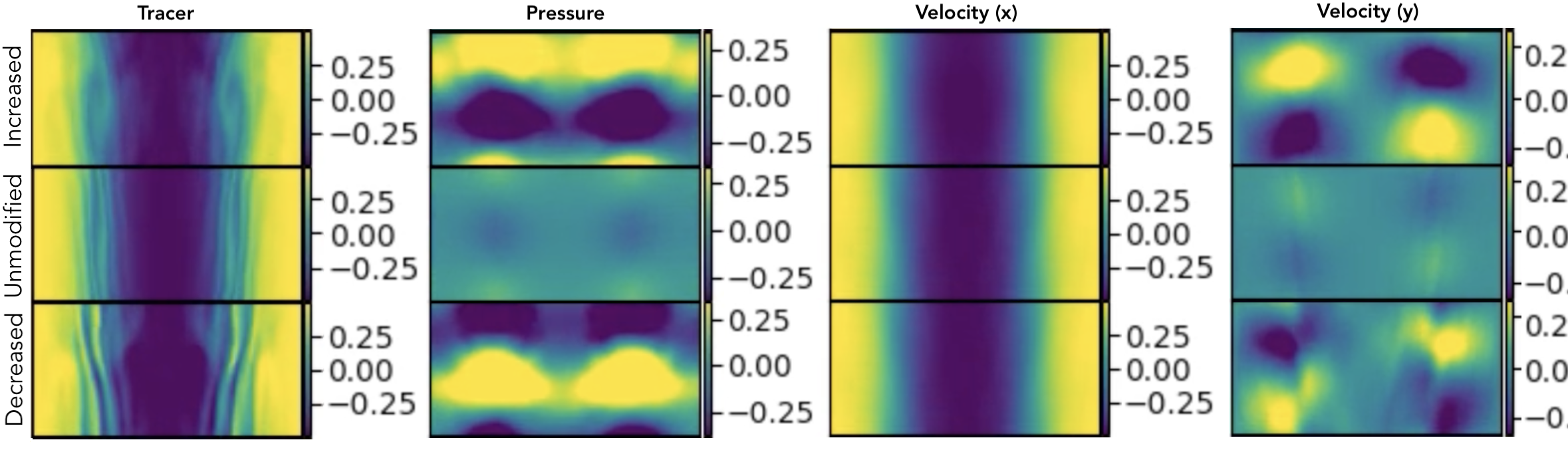}
    \caption{Trace, pressure, x velocity and y velocity for $\scriptstyle\CDelta_{\text{vortex}}$ injection into shear flow laminar regime with (top) $\alpha=\mathit{0.2}$, (middle) $\alpha=\mathit{0.0}$ and (bottom) $\alpha=\mathit{-0.2}$. Frame: 64.}
    \label{fig:vortex_fields}
\end{figure}

\begin{figure}[H]
    \centering
    \includegraphics[width=0.8\linewidth]{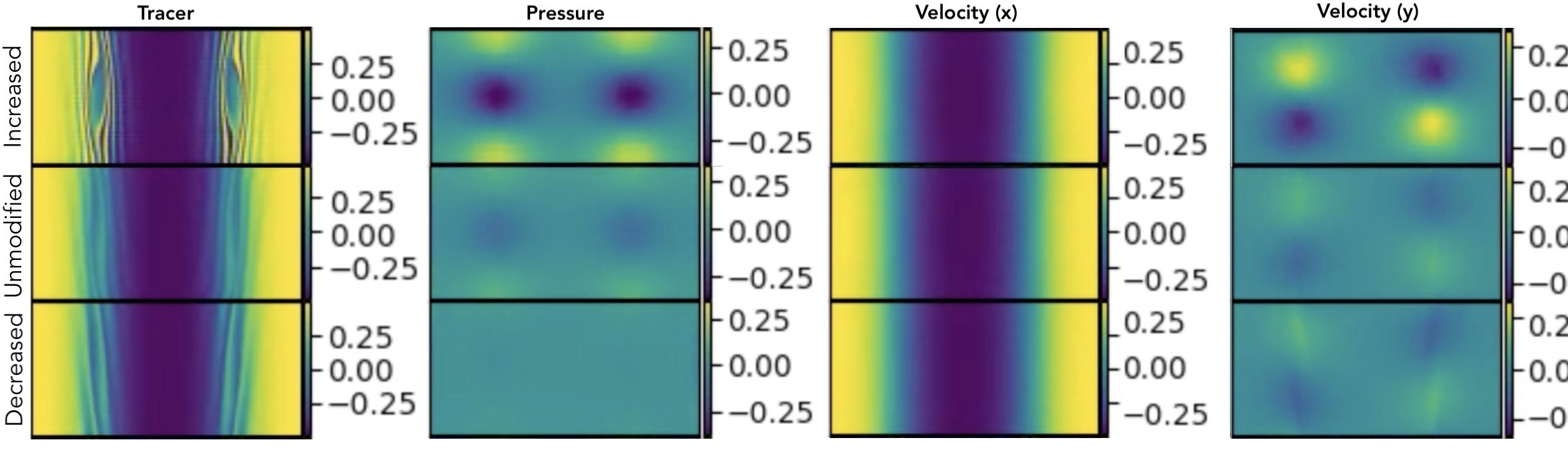}
    \caption{Trace, pressure, x velocity and y velocity for $\scriptstyle\overline{\CDelta}_{\text{vortex}}$ injection into shear flow laminar regime with (top) $\alpha=\mathit{0.7}$, (middle) $\alpha=\mathit{0.0}$ and (bottom) $\alpha=\mathit{-0.5}$. Frame: 64.}
    \label{fig:vortex_fields_drop}
\end{figure}

In \cref{fig:vortex_fields} we visualise the effect of shear flow vorticity steering with spatial dimensions on related physical fields beyond the primary tracer field. It is clear that the activation steering does not merely alter the tracer output, but introduces coordinated changes across the pressure and velocity fields that on the surface seem consistent with real vortex dynamics. Coordinated changes which evolve and persist throughout the rollout window (frame 64 being the final frame of the rollout window). These modifications are bidirectional, with positive and negative steering causing opposite changes to the pressure and y velocity fields. It should be noted that this result is not consistent with the interpretation of a linear steering response, where positive steering amplifies the concept feature and negative steering suppresses it. This is because while the positive steering result looks as one might expect, the negative steering result does not, since if the negative steering was causing a decrease in vorticity we would expect a flattening of the pressure field, not an inversion.

In \cref{fig:vortex_fields_drop} we visualise the effect of shear flow vorticity steering \emph{without} spatial dimensions. Two key differences can be see with the previous \cref{fig:vortex_fields}. Firstly, the primary tracer displays a smaller, less well-formed vortex in the positive direction (despite a higher value of $\alpha$ than \cref{fig:vortex_fields}) and the negative direction displays a more natural-looking laminar flow. Secondly, the changes to the secondary fields are more subtle in the positive direction and no longer inverted in the negative direction. A result which \emph{is} consistent with the interpretation of a linear steering response.

These results are comparable to the Rayleigh-Bénard results in \cref{fig:ray-transfer}, and other experiments we have performed:
\begin{itemize}
    \item When spatial dimensions are averaged ($\overline{\CDelta}_f$) over results are consistent with the interpretation of a linear steering response.
    \item When this does not occur and spatial dimensions are left in place ($\CDelta_f$) steering has a tendency to induce mirrored field changes in positive and negative steering. 
    \item $\overline{\CDelta}_f$ steering usually produces more natural-looking results.
    \item $\CDelta_f$ steering, on the other hand, has the capability to produce more extreme changes, such as the creation of large, intricate vorticities, but at the expense of a less natural-looking final result across all fields. 
\end{itemize}

\subsection{Shear Flow Diffusion Steering – Additional Fields}
\begin{figure}[H]
    \centering
    \includegraphics[width=0.8\linewidth]{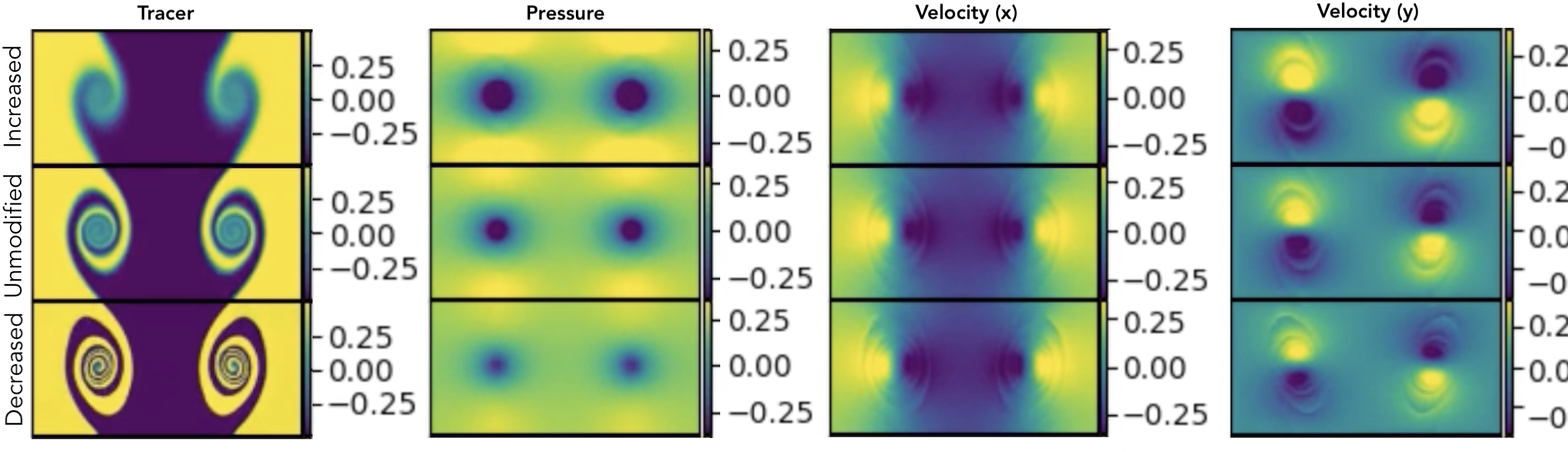}
    \caption{Trace, pressure, x velocity and y velocity for $\scriptstyle\CDelta_{\text{diffusion}}$ injection into shear flow vortex regime with (top) $\alpha=\mathit{0.2}$, (middle) $\alpha=\mathit{0.0}$ and (bottom) $\alpha=\mathit{-0.2}$. Frame: 28.}
    \label{fig:diffusion_fields}
\end{figure}

\section{Data Files}
\subsection{Simulations}
\begin{table}[H]
    \centering
    \caption{Simulation Data Files}
  \begin{tabular}{cc}
    \toprule
    Simulation & Datafile \\
    \midrule
    \cref{fig:vortex_suppress_shear} (all) & shear\_flow\_Reynolds\_5e4\_Schmidt\_2e-1 \\
    \cref{fig:vortex_addition_shear} (all) & shear\_flow\_Reynolds\_5e5\_Schmidt\_1e0 \\
    \cref{fig:shear_additional_concepts} (left) & shear\_flow\_Reynolds\_5e4\_Schmidt\_2e-1 \\
    \cref{fig:shear_additional_concepts} (right) & shear\_flow\_Reynolds\_5e4\_Schmidt\_5e-1 \\
    \cref{fig:ray-transfer} (all) & rayleigh\_bernard\_rayleigh\_1e9\_prandtl\_10 \\
    \cref{fig:euler-gray-transfer} (left) & euler\_multi\_quadrants\_openBC\_gamma\_1.33\_H2O\_20 \\
    \cref{fig:euler-gray-transfer} (middle) & euler\_multi\_quadrants\_openBC\_gamma\_1.404\_H2\_100\_Dry\_air\_-15 \\
    \cref{fig:euler-gray-transfer} (right) & gray\_scott\_reaction\_diffusion\_gliders\_F\_0.014\_k\_0.054 \\
    \cref{fig:vortex_fields} (all) & shear\_flow\_Reynolds\_1e5\_Schmidt\_2e-1 \\
    \cref{fig:vortex_fields_drop} (all) & shear\_flow\_Reynolds\_1e5\_Schmidt\_2e-1 \\
    \cref{fig:diffusion_fields} (all) & shear\_flow\_Reynolds\_5e4\_Schmidt\_2e-1 \\
    \bottomrule
  \end{tabular}
\end{table}

\subsection{Steering Tensors}
\begin{table}[H]
\begin{minipage}{.48\linewidth}
    \centering
    \caption{Vortex Regime Simulations}
        \begin{tabular}{ccc}
          \toprule
          Name & Reynolds & Schmidt \\
          \midrule
          Shear Flow & 1e4 & 1e-1 \\
          Shear Flow & 1e4 & 2e-1 \\
          Shear Flow & 1e4 & 2e0 \\
          Shear Flow & 1e4 & 5e-1 \\
          Shear Flow & 1e4 & 5e0 \\
          Shear Flow & 1e5 & 1e-1 \\
          Shear Flow & 1e5 & 1e0 \\
          Shear Flow & 1e5 & 2e0 \\
          Shear Flow & 1e5 & 5e-1 \\
          Shear Flow & 5e4 & 1e-1 \\
          Shear Flow & 5e4 & 1e0 \\
          Shear Flow & 5e4 & 1e1 \\
          Shear Flow & 5e4 & 2e0 \\
          Shear Flow & 5e4 & 5e-1 \\
          Shear Flow & 5e4 & 5e0 \\
          Shear Flow & 5e5 & 1e0 \\
          Shear Flow & 5e5 & 2e-1 \\
          Shear Flow & 5e5 & 5e0 \\
          \bottomrule
        \end{tabular}
\end{minipage}
\begin{minipage}{.5\linewidth}
    \centering
    \caption{Laminar Regime Simulations}
        \begin{tabular}{ccc}
          \toprule
          Name & Reynolds & Schmidt \\
          \midrule
          Shear Flow & 1e4 & 1e0 \\
          Shear Flow & 1e4 & 1e1 \\
          Shear Flow & 1e5 & 1e1 \\
          Shear Flow & 1e5 & 2e-1 \\
          Shear Flow & 1e5 & 5e0 \\
          Shear Flow & 5e4 & 2e-1 \\
          Shear Flow & 5e5 & 1e-1 \\
          Shear Flow & 5e5 & 1e1 \\
          Shear Flow & 5e5 & 2e0 \\
          Shear Flow & 5e5 & 5e-1
        \end{tabular}
\end{minipage}
\end{table}

\begin{table}[H]
    \centering
    \caption{High Diffusion Simulation}
  \begin{tabular}{cccccc}
    \toprule
    Name & Regime & Reynolds & Schmidt & Viscosity & Diffusion \\
    \midrule
    Shear Flow & single vortex & 5e4 & 2e-1 & 2.00e-05 & 1.00e-04 \\
    \bottomrule
  \end{tabular}
\end{table}
\begin{table}[H]
    \centering
    \caption{Low Diffusion Simulation}
  \begin{tabular}{cccccc}
    \toprule
    Name & Regime & Reynolds & Schmidt & Viscosity & Diffusion \\
    \midrule
    Shear Flow & single vortex & 5e4 & 1e1 & 2.00e-05 & 2.00e-06 \\
    \bottomrule
  \end{tabular}
\end{table}

\begin{table}[H]
    \centering
    \caption{High Speed Simulations}
      \begin{tabular}{cccc}
      \toprule
      Name & Reynolds & Schmidt & dt\_stride \\
      \midrule
      Shear Flow & 1e4 & 1e0 & 2 \\
      Shear Flow & 1e4 & 1e1 & 2 \\
      Shear Flow & 1e5 & 1e1 & 2 \\
      Shear Flow & 1e5 & 2e-1 & 2 \\
      Shear Flow & 1e5 & 5e0 & 2 \\
      Shear Flow & 5e4 & 2e-1 & 2 \\
      Shear Flow & 5e5 & 1e-1 & 2 \\
      Shear Flow & 5e5 & 1e1 & 2 \\
      Shear Flow & 5e5 & 2e0 & 2 \\
      Shear Flow & 5e5 & 5e-1 & 2 \\
      \bottomrule
      \end{tabular}
\end{table}
\begin{table}[H]
    \centering
    \caption{Low Speed Simulations}
        \begin{tabular}{cccc}
        \toprule
        Name & Reynolds & Schmidt & dt\_stride \\
        \midrule
        Shear Flow & 1e4 & 1e0 & 1 \\
        Shear Flow & 1e4 & 1e1 & 1 \\
        Shear Flow & 1e5 & 1e1 & 1 \\
        Shear Flow & 1e5 & 2e-1 & 1 \\
        Shear Flow & 1e5 & 5e0 & 1 \\
        Shear Flow & 5e4 & 2e-1 & 1 \\
        Shear Flow & 5e5 & 1e-1 & 1 \\
        Shear Flow & 5e5 & 1e1 & 1 \\
        Shear Flow & 5e5 & 2e0 & 1 \\
        Shear Flow & 5e5 & 5e-1 & 1 \\
        \bottomrule
        \end{tabular}
\end{table}

\end{document}